\documentclass{article}

\usepackage[preprint]{neurips_2025}

\usepackage[utf8]{inputenc} %
\usepackage[T1]{fontenc}    %
\usepackage{hyperref}       %
\usepackage{url}            %
\usepackage{booktabs}       %
\usepackage{amsfonts}       %
\usepackage{nicefrac}       %
\usepackage{microtype}      %
\usepackage{xcolor}         %

\usepackage{graphicx} %
\usepackage{natbib}  %
\usepackage{caption} %
\usepackage{algorithm}
\usepackage{listings}
\usepackage{amsmath}
\usepackage{multirow}
\usepackage[outdir=./]{epstopdf}
\usepackage{enumitem}
\usepackage{caption}
\usepackage{subcaption}
\usepackage{subfloat}
\usepackage{newfloat}
\usepackage{graphicx}
\usepackage{svg} %
\usepackage[normalem]{ulem}
\usepackage{framed}
\usepackage{mdframed}
\usepackage{lipsum}
\usepackage{float}
\usepackage{xurl}
\usepackage{makecell}
\usepackage{array}
\usepackage{multirow}
\usepackage{wrapfig}
\usepackage{makecell}
\usepackage{enumitem}
\usepackage{pifont}
\newcommand{\xmark}{\ding{55}} %
\usepackage{algpseudocode}
\usepackage{amsmath, amssymb}
\usepackage{setspace}
\usepackage{float}
\usepackage[most]{tcolorbox}
\renewcommand{\thefootnote}{\fnsymbol{footnote}}

\usepackage{colortbl}
\usepackage[table]{xcolor} %
\definecolor{lightgray}{rgb}{0.9, 0.9, 0.9}
\definecolor{lightblue}{rgb}{0.8, 0.9, 1.0}
\definecolor{lightgreen}{rgb}{0.8, 1.0, 0.8}
\definecolor{taskgreen}{RGB}{232,245,233}
\definecolor{taskblue}{RGB}{232,240,254}
\definecolor{taskorange}{RGB}{253,237,228}

\makeatletter
\@ifundefined{insightbox}{
  \@ifundefined{newtcolorbox}{
    \newenvironment{insightbox}[1]{\begin{quote}\noindent\textbf{#1. }\itshape}{\end{quote}}
  }{
    \newtcolorbox{insightbox}[2][]{
      enhanced,
      breakable,
      colback=blue!3!white,
      colframe=blue!55!black,
      boxrule=0.6pt,
      arc=2mm,
      left=1.2mm,right=1.2mm,top=1mm,bottom=1mm,
      title={#2},
      fonttitle=\bfseries,
      #1
    }
  }
}{}
\makeatother
\title{SpatialAct: Probing Spatial Reasoning-to-Action Capabilities of VLM Agents in 3D Scenes}

\author{
\centerline{
\bf Tianhui Liu$^{1}$ \quad Jie Feng$^{2\dagger}$ \quad Zhiheng Zheng$^3$ \quad Shengyuan Wang$^3$ } \\
\centerline{
\bf Yiming Guo$^2$ \quad Yanxin Xi$^4$ \quad Hangyu Fan$^3$ \quad Yong Li$^{3\dagger}$ \quad Pan Hui$^{1,4\dagger}$
} \\ \\
$^{1}$The Hong Kong University of Science and Technology (Guangzhou) 
\\ $^2$Zhongguancun Academy \\
$^3$Tsinghua University
\\ $^4$Helsinki University
}

\begin{document}

\maketitle

\begin{abstract}
  Humans can effortlessly perceive spatial layouts, form cognitive representations, reason about spatial relations, and translate such reasoning into actions in everyday 3D environments. Although recent vision-language models (VLMs) have shown promising performance on observation-conditioned spatial perception and reasoning tasks, it remains unclear whether they can build coherent spatial understanding, act upon it, and refine their actions through multi-turn feedback. To study this problem, we introduce \textbf{SpatialAct}, a simulator-grounded benchmark for probing \textit{action-conditioned spatial reasoning} in 3D scenes. Starting from the most challenging setting, Multi-turn Interactive Refinement, we further design its decomposed counterpart, Single-step Error Detection and Fix, together with five fundamental spatial ability tasks to diagnose the underlying causes of model failures. Experiments reveal a clear reasoning-to-action gap: current VLMs can perform well on isolated spatial reasoning tasks, but struggle to maintain coherent spatial beliefs and produce reliable actions during multi-turn feedback, substantially underperforming humans. These results suggest that current VLM agents still lack robust spatial state tracking under action-induced environment changes, even when low-level control is abstracted away.
\end{abstract}

\setcounter{footnote}{0} %
\footnotetext[2]{Corresponding author, email: fengj12ee@hotmail.com, liyong07@tsinghua.edu.cn, panhui@ust.hk.}
\renewcommand{\thefootnote}{\arabic{footnote}}
\setcounter{footnote}{0}

\section{Introduction} \label{sec:introduction}

Vision-language models are increasingly moving from passive perception toward agentic interaction in 3D environments and the real world~\citep{majumdar2024openeqa,wang2025embodied,zhao2025cityeqa,zhang2025etplan}. This shift is driven by emerging applications such as indoor scene understanding and editing, embodied manipulation, 3D world generation, and interactive simulation~\citep{abdelreheem2025placeit3d,el2025scanedit,feng2023layoutgpt,sun2025layoutvlm,weihs2021rearrangement,yang2024holodeck,zhang2025etplan,zhen20263d-layout-r1}. In these settings, spatial intelligence is no longer limited to recognizing objects or answering questions about spatial relations. A spatially capable agent must build and maintain a coherent understanding of the environment, decide how to intervene, observe how the environment changes after its own action, and continue reasoning over the updated state. This ability is fundamental for bridging visual-spatial understanding with real-world and simulator-grounded agency.

Recent spatial reasoning benchmarks have made substantial progress in evaluating the visual-spatial capabilities of VLMs. Existing benchmarks examine spatial understanding from different perspectives, including 2D spatial relations, classical spatial cognition, 3D spatial reasoning, and psychometric-inspired basic spatial abilities~\citep{huang2025spatial,li2024core,wang2024picture,wang2025site,wang2025spatial457,xu2025defining,zhang-etal-2025-sphere}. Other efforts further extend spatial evaluation to richer and more realistic settings, such as all-scale spatial reasoning from millimeters to kilometers~\citep{sun2025spacevista}, 3D scenes and videos~\citep{yang2025thinking}, limited-view spatial mental modeling~\citep{yin2025spatial}, cross-view urban reasoning~\citep{xu2026citycube}, embodied question answering~\citep{majumdar2024openeqa,zhao2025cityeqa}, and active spatial belief construction~\citep{zhang2026theory}. These benchmarks have revealed important limitations of current VLMs in perceiving, remembering, and reasoning about space.

Nevertheless, existing evaluations still leave an important gap. As summarized in Table~\ref{tab:benchmark_comparison}, most spatial reasoning benchmarks evaluate models as observers. The model is given an image, a video, or a set of views, and is asked to answer questions about the observed scene. Even when the input becomes multi-view, egocentric, or temporally extended, the model output usually does not directly alter the environment that the model must subsequently reason about. In contrast, embodied benchmarks involve actions and feedback, but their evaluation often entangles high-level spatial reasoning with low-level control, navigation, or manipulation~\citep{wang2025embodied,weihs2021rearrangement,zhang2025etplan}. Therefore, a missing middle ground remains between passive spatial question answering and full embodied control: evaluating whether VLM agents can perform high-level semantic spatial actions, observe the resulting state transitions, and maintain consistent reasoning across multiple rounds of interaction.

\begin{table*}[t]
\centering
\caption{Comparison of spatial reasoning and embodied benchmarks.}
\label{tab:benchmark_comparison}
\resizebox{\textwidth}{!}{
\setlength{\tabcolsep}{1pt} %
\begin{tabular}{lccccccccc}
\toprule
\textbf{Benchmark} & \textbf{Dimension} & \textbf{Modality} & \textbf{Scenario} & \textbf{QA Size} & \textbf{Eval Type} & \textbf{Multi-view}  &\textbf{Multi-turn} & \textbf{Env. Feed.} & \textbf{Objective} \\
\midrule
SpatialEval~\cite{wang2024picture}& 2D  & Image & Mix     & 4.6k  & MCQ        & \xmark  &\xmark     & \xmark     & Understanding \\
CoreCognition~\cite{li2024core}   & 2D  & Mix   & Mix     & 1.5k  & MCQ        & \xmark  &\xmark     & \xmark     & Understanding \\
BSA~\cite{xu2025defining}             & Mix & Image & Classic & 312   & MCQ        & \xmark  &\xmark     & \xmark     & Understanding \\
Spatial457~\cite{wang2025spatial457}      & Mix & Image & Outdoor & 23.7k & Mix        & \xmark  &\xmark     & \xmark     & Understanding \\
VIS-Bench~\cite{yang2025thinking}       & 3D  & Video & Indoor  & 5k    & Mix        & \checkmark  &\xmark  & \xmark     & Understanding \\
MINDCUBE~\cite{yin2025spatial}        & 3D  & Image & Mix     & 21.1k & MCQ        & \checkmark  &\xmark  & \xmark     & Understanding \\
CityCube~\cite{xu2026citycube}        & 3D  & Image & Outdoor & 5.0k  & MCQ        & \checkmark  &\xmark  & \xmark     & Understanding \\
OpenEQA~\cite{majumdar2024openeqa}         & 3D  & Video & Indoor  & 1.6k  & Open-ended & \checkmark  &\xmark  & \xmark     & Exploration \\
THEORY OF SPACE~\cite{zhang2026theory} & 3D  & Image & Indoor  & 2.7k  & Open-ended & \checkmark  &\checkmark & \xmark  & Exploration \\
MetaVQA~\cite{wang2025embodied}         & 3D  & Image & Outdoor & 9.7k  & MCQ        & \xmark  &\checkmark & \checkmark & Action \\
\midrule
\textbf{SpatialAct}      & 3D  & Image & Mix     & 4.3k  & Mix        & \checkmark  &\checkmark & \checkmark & Action \\
\bottomrule
\end{tabular}
}
\vspace{-25pt}
\end{table*}

We refer to this capability as \textbf{action-conditioned spatial reasoning}. Unlike conventional observation-conditioned reasoning, action-conditioned spatial reasoning requires a model to reason not only about the current spatial state, but also about how its own action changes that state and how future decisions should adapt to the updated environment. To make this evaluation concrete and controllable, we instantiate action-conditioned spatial reasoning as interactive 3D layout refinement. This formulation is motivated by the fact that spatial validity, including collision avoidance, boundary consistency, and orientation plausibility, is a fundamental requirement for usable 3D scenes~\citep{abdelreheem2025placeit3d,el2025scanedit,feng2026repurposing,feng2023layoutgpt,huang2025fireplace,sun2025layoutvlm,yang2024holodeck,zhen20263d-layout-r1}. At the same time, layout refinement provides executable high-level actions and objectively verifiable state changes, allowing us to isolate spatial reasoning from low-level motor control.

To this end, we introduce \textbf{SpatialAct}, a simulator-grounded benchmark for probing reasoning-to-action capabilities of VLM agents in 3D scenes. \textbf{SpatialAct} abstracts away low-level control and focuses on high-level semantic spatial actions, where models issue executable commands such as moving, rotating, or scaling objects with specific parameters. These commands are parsed and executed in a 3D simulator, and the updated multi-view renderings are then returned to the model for continued reasoning and action. \textbf{SpatialAct} covers three types of 3D scenarios, including \textit{Abstract Geometric}, \textit{Urban Architectural}, and \textit{Indoor Scene}, with 333 scenes and 4,355 QA pairs across three question formats. Each scene is rendered from both top-view and isometric-view perspectives, while procedural generation, simulator execution, and dynamically injected spatial errors enable controllable evaluation and reduce the risk of static benchmark contamination.

\textbf{SpatialAct} follows a hierarchical diagnostic design that connects basic spatial understanding to multi-turn action-conditioned refinement. It first evaluates five Basic Spatial Abilities, including object meaning, spatial relations, spatial orientation, mental rotation, and spatial visualization. It then tests whether models can identify and repair spatial errors in one step through Single-step Error Detection and Fix. Finally, it challenges models to iteratively repair abnormal spatial configurations through closed-loop simulator feedback in Multi-turn Interactive Refinement. Empirically, our experiments reveal a substantial reasoning-to-action gap. Although current strong VLMs perform well on isolated spatial tasks and reach around 80\% accuracy in several basic categories, the strongest VLM only achieves 0.411 Repair Rate and 0.206 Scene Success Rate in multi-turn simulator-grounded refinement. In contrast, human participants achieve 0.911 Repair Rate and 0.763 Scene Success Rate. These results suggest that current VLMs may recognize local spatial relations, but still struggle to maintain coherent spatial beliefs and produce reliable actions across long-horizon state transitions. Our contributions are threefold.
\begin{itemize}[leftmargin=1.5em,itemsep=0pt,parsep=0.2em,topsep=0.0em,partopsep=0.0em]
\item We formalize \textbf{action-conditioned spatial reasoning} as a missing evaluation axis for VLM agents, where high-level simulator-executable actions change the 3D environment state that future reasoning depends on.
\item We introduce \textbf{SpatialAct}, a simulator-grounded benchmark with 333 scenes and 4,355 QA pairs across \textit{Abstract Geometric}, \textit{Urban Architectural}, and \textit{Indoor Scene} scenarios, organized into a hierarchical design covering Basic Spatial Abilities, Single-step Error Detection and Fix, and Multi-turn Interactive Refinement.
\item We conduct a systematic evaluation of leading proprietary and open-source VLMs, revealing a clear reasoning-to-action gap between basic spatial understanding and stable multi-turn spatial refinement. We will release the benchmark, simulator workflow, evaluation platform, and analysis toolkit to support future research on spatially grounded VLM agents. Our codes and datasets are open-sourced via \url{https://github.com/tsinghua-fib-lab/SpatialAct}.
\end{itemize}

\vspace{-5pt}
\section{Methods} \label{sec:methods}
\vspace{-5pt}
\begin{figure}[t]
    \centering
    \includegraphics[width=0.99\linewidth]{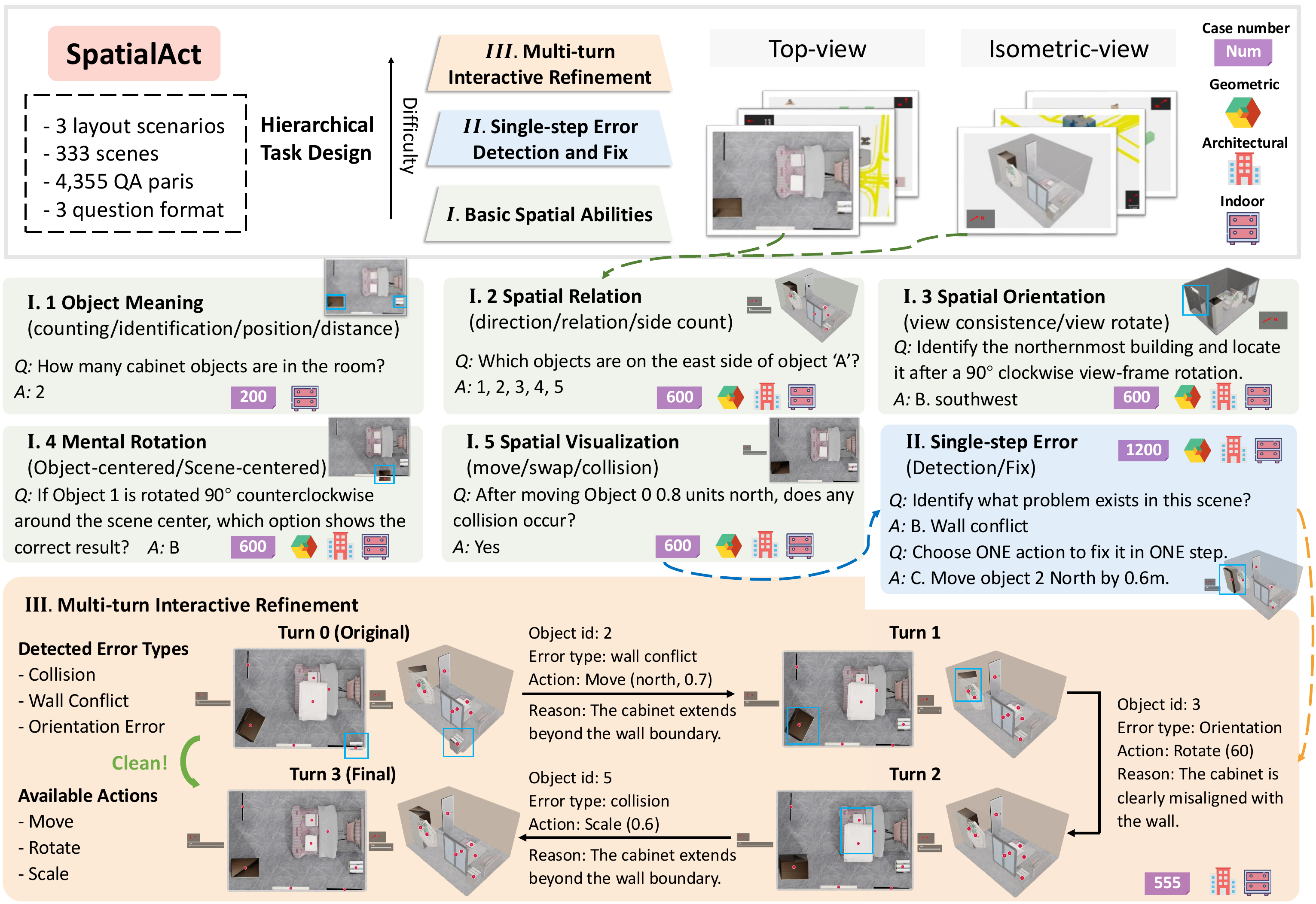}
    \caption{Framework of SpatialAct, which introduces a three-level hierarchical task design to systematically evaluate and diagnose agentic VLMs' 3D spatial reasoning and action capabilities.}
    \label{fig:framework}
    \vspace{-15pt}
\end{figure}
We introduce \textbf{SpatialAct} to evaluate agentic VLMs in 3D environments from reasoning to action. As shown in Figure~\ref{fig:framework}, SpatialAct covers three types of layout scenarios: Abstract Geometric, Urban Architectural, and Indoor Scene, where the latter two are constructed from 333 Scenes. In total, SpatialAct contains 4,355 QA pairs spanning 3 question formats, providing a unified testbed for evaluating spatial understanding across controlled geometric layouts and daily-life 3D scenes.
\vspace{-5pt}
\subsection{Benchmark Construction}
\begin{figure}[t]
    \centering
    \includegraphics[width=0.99\linewidth]{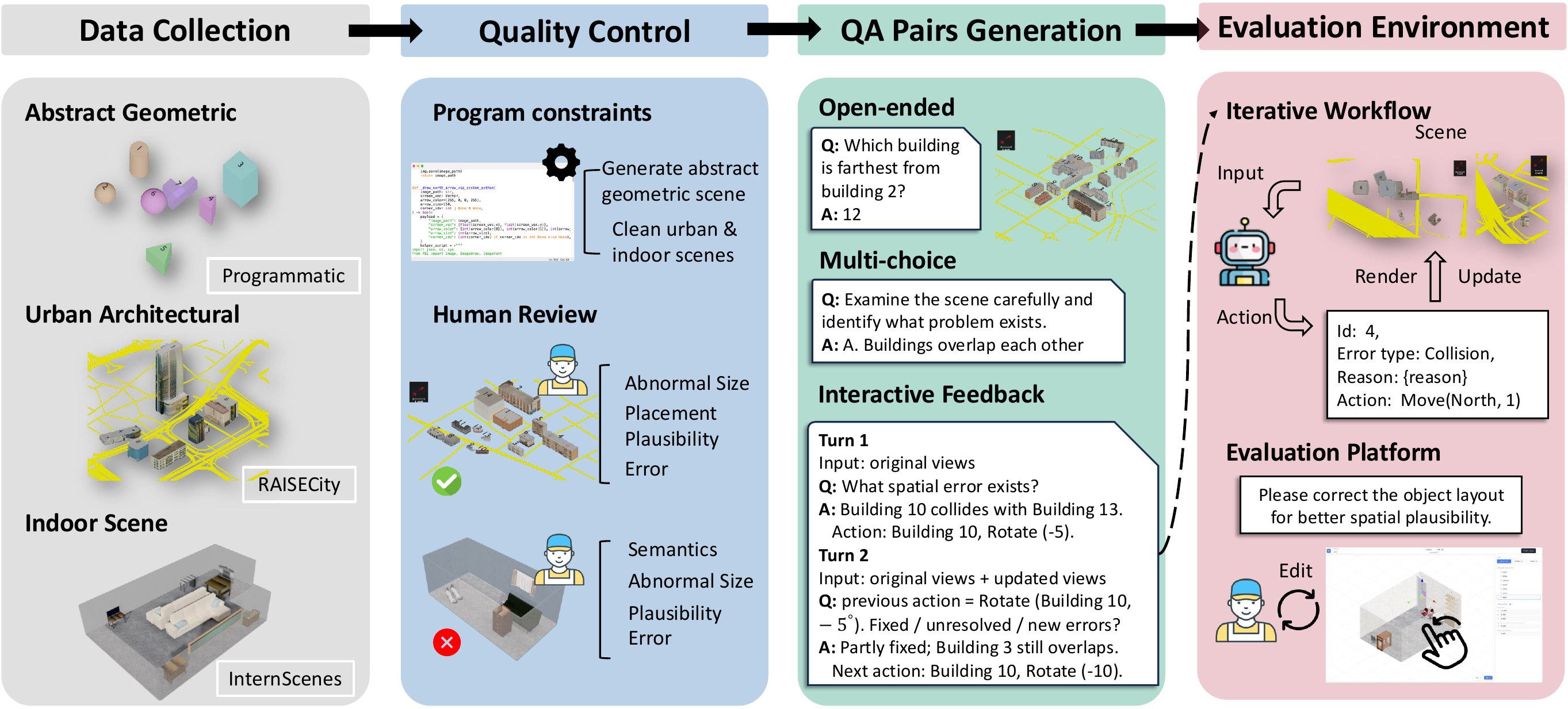}
    \caption{Benchmark construction pipeline, including data collection, quality control, QA pair generation, and evaluation environment setup.}
    \label{fig:pipeline}
    \vspace{-15pt}
\end{figure}

\vspace{-5pt}
To evaluate the 3D spatial reasoning and action capabilities of VLMs across diverse scenarios, we construct a scene-level dataset as illustrated in Figure~\ref{fig:pipeline}. The pipeline consists of data collection, quality control, and QA pair generation. Each scene is represented by both a top-view and an isometric-view, enabling models to access complementary spatial information from different perspectives.

\vspace{-5pt}
\paragraph{Data Collection}
\vspace{-5pt}
SpatialAct consists of three types of scenarios: Abstract Geometric, Urban Architectural, and Indoor Scene. The Abstract Geometric layouts are procedurally generated in a controllable manner. Each scene typically contains 5-7 geometric objects, which are randomly sampled from a predefined set of candidates, including cube, cuboid, cylinder, prism, sphere, L-shape, and U-shape, with varying sizes. For more complex tasks such as Mental Rotation, we further constrain each scene to contain at least one complex L-shaped or U-shaped object, ensuring sufficient structural complexity for spatial reasoning.
The Urban Architectural layouts are derived from RAISECity~\cite{wang2025raisecity}, a multimodal agent framework for reality-aligned 3D world generation. We use both its white-box building models and textured building models to construct mixed urban scenes, while restricting each scene to contain no more than 20 buildings. The Indoor Scene layouts are built from InternScenes~\cite{zhong2025internscenes}, which provides diverse indoor environments with movable furniture meshes. Considering the higher visual and structural complexity of indoor environments, we select scenes containing 5--15 objects for benchmark construction.
\vspace{-10pt}
\paragraph{Quality Control}
For Abstract Geometric scenes, the layouts are generated procedurally under predefined constraints, and therefore no additional filtering is required. For Urban Architectural and Indoor Scene layouts, although the original scene-generation pipelines already provide relatively well-structured scenes, our benchmark focuses on error patterns in spatial configurations. We therefore first apply programmatic cleaning to remove existing abnormal cases from the original scenes.
We then conduct manual filtering to examine abnormal element sizes, the plausibility of spatial arrangements, and potential error patterns. For Indoor Scene layouts, we additionally check semantic consistency, as indoor objects carry rich semantic categories. These inspections ensure the quality and reliability of the final benchmark scenes.
\vspace{-10pt}
\paragraph{Task Design and QA Pairs Generation}
SpatialAct follows a hierarchical task design centered on Multi-turn Interactive Refinement, the most challenging setting in our benchmark. In this task, we first inject layout errors of varying difficulty and correction steps into clean, error-free scenes obtained from the previous filtering stage. Models are required to iteratively repair abnormal errors in a scene through multi-turn action-based interaction. To facilitate analysis of the underlying causes of model performance, we further design two levels of diagnostic tasks. First, Single-step Error Detection and Fix decomposes the interactive refinement process into a one-step setting, focusing on whether models can detect a spatial error and choose the corresponding correction. Second, we introduce five basic spatial ability tasks: Object Meaning, Spatial Relation, Spatial Orientation, Mental Rotation, and Spatial Visualization. These dimensions are adapted from prior work on basic spatial abilities~\cite{xu2025defining}, which identifies five key dimensions based on psychometric theories. While the original tasks are mainly designed around classical geometry and cube-like scenarios, we reinterpret their theoretical meanings and redesign the questions for Abstract Geometric scenes, further extending them to more common daily-life scenarios, including Urban Architectural and Indoor Scene environments. Each task category contains multiple QA subtypes, each emphasizing different aspects of spatial cognition. Detailed QA subtype definitions are provided in the Appendix~\ref{app:subcate_bsa}. Depending on the characteristics of each task, QA pairs are formulated in three formats: open-ended questions, multiple-choice questions (MCQs), and multi-turn feedback-based interaction.

\vspace{-5pt}
\subsection{Evaluation Environment Setup}
\begin{figure}[t]
    \centering
    \includegraphics[width=0.99\linewidth]{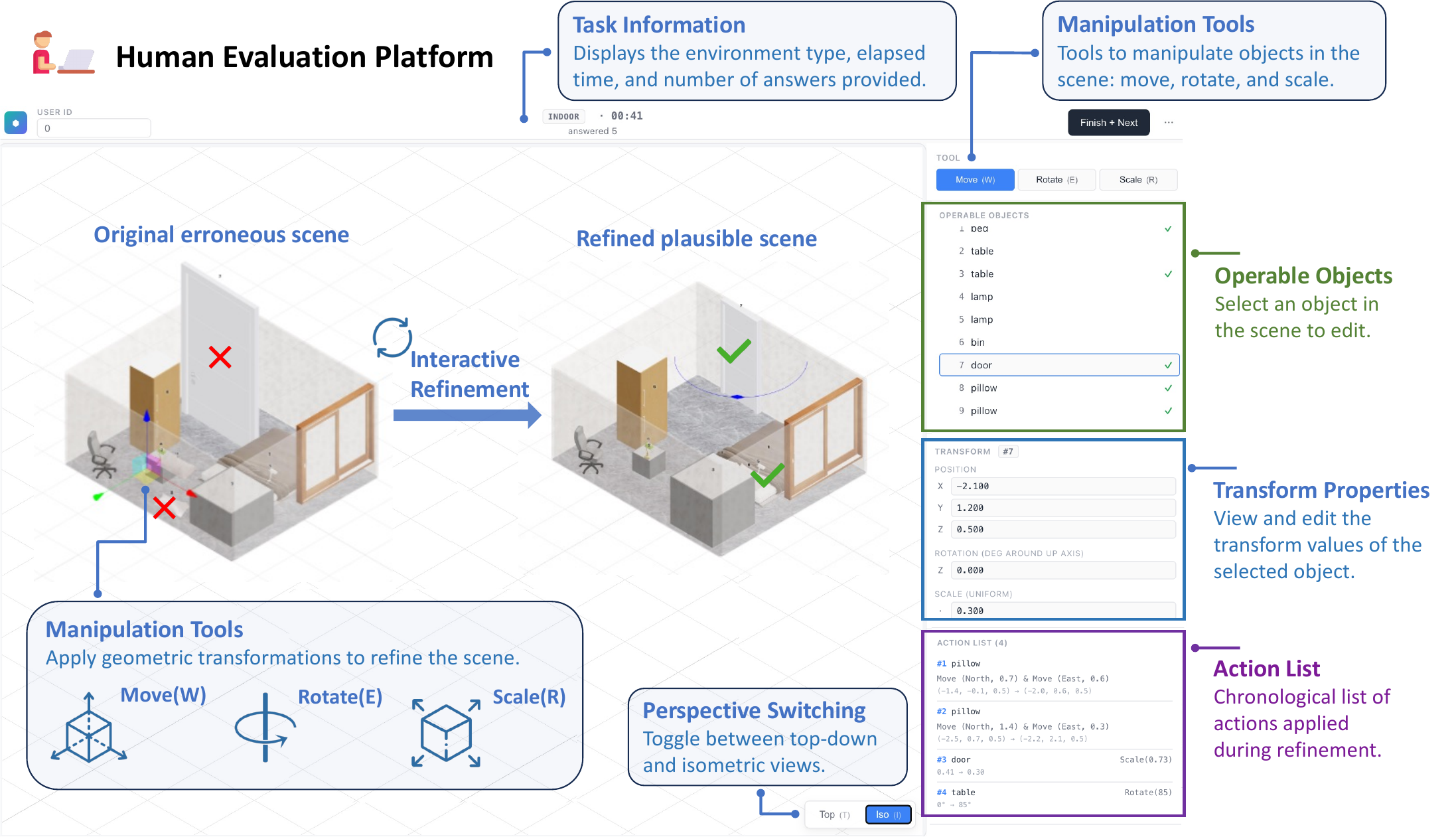}
    \caption{The web-based evaluation interface for interactive scene refinement.}
    \label{fig:evaluation-platform}
    \vspace{-15pt}
\end{figure}

\vspace{-5pt}
To evaluate the reasoning-to-action capabilities of current VLMs, we design the Multi-turn Interactive Refinement task, where models are required to interact with the environment through iterative action-based feedback. Unlike static QA tasks, this setting requires the model to inspect the current scene, identify spatial errors, generate corrective actions, and refine its decisions based on the updated environment. In the following, we describe its implementation through the Multi-turn Iterative Rendering Workflow and the human Evaluation Platform.
\vspace{-10pt}
\paragraph{Multi-turn Iterative Rendering Workflow}
In the Multi-turn Interactive Refinement task, VLMs are required to detect and correct abnormal spatial configurations through iterative action-based interaction. These error types are designed to reflect common failure modes in 3D layout construction~\cite{sun2025layoutvlm,feng2026repurposing}, where objects or buildings must avoid physical conflicts, stay within valid spatial boundaries, and maintain plausible orientations. For Urban Architectural scenes, the model is asked to identify three types of building-related errors: collision, road conflict, and orientation error. For Indoor Scene layouts, the model focuses on three types of object-related errors: collision, wall conflict, and orientation error. At each turn, the model can manipulate the current scene using three types of actions: \texttt{move(direction, number)}, \texttt{rotate(degree)}, and \texttt{scale(number)}.
As illustrated in the rightmost part of Figure~\ref{fig:pipeline}, each interaction starts by providing the VLM with the top-view and isometric-view renderings of an original erroneous scene. The model then determines which spatial errors exist and outputs corrective action commands. These commands are parsed by and executed in the simulator, after which the updated scene is rendered again from both the top and isometric views. In the next turn, the VLM receives the newly rendered views together with the original top-view and isometric-view images, the last-turn action command, and the accumulated previous action history. This iterative process continues until the VLM judges that all errors in the scene have been fixed or the predefined maximum number of turns is reached.
\vspace{-10pt}
\paragraph{Human Evaluation Platform}
To enable human participants to perform the Multi-turn Interactive Refinement task, we develop a web-based evaluation interface. As illustrated in Figure~\ref{fig:evaluation-platform}, the platform provides an intuitive 3D environment equipped with a Transform Gizmo, allowing participants to perform free-form editing by directly manipulating objects in the scene. Users can select target entities from the Object Checklist and precisely adjust their spatial properties (position, rotation, and scale) using either the interactive handle or the Transform Properties panel. To ensure high-quality refinement, the interface also features perspective switching for multi-angle inspection and an Operation History log for tracking iterative modifications. This flexible workflow empowers human users to effectively rectify spatial inconsistencies, ensuring the final generated scenes are both structurally and functionally plausible.

\begin{table*}[t]
\centering
\caption{Main results on the SpatialAct. Rep. Rate denotes Repair Rate, Succ. Rate denotes Scene Success Rate, Eff. Turn denotes Effective Repair Turn Ratio, Pre. Stop denotes Premature Stop Ratio, and Tok./Scene denotes Average Completion Tokens per Scene. }
\label{tab:main_results}
\resizebox{\textwidth}{!}{
\begin{tabular}{lccccccccccc}
\toprule
\multirow{2}{*}{\makecell{\textbf{Task}}} 
& \multicolumn{5}{>{\columncolor{taskorange}}c}{\textbf{Multi-turn Refinement}}
& \cellcolor{taskblue}\makecell{\textbf{Single-step}\\\textbf{Edit}}
& \cellcolor{taskgreen}\makecell{\textbf{Object}\\\textbf{Meaning}}
& \cellcolor{taskgreen}\makecell{\textbf{Spatial}\\\textbf{Relation}}
& \cellcolor{taskgreen}\makecell{\textbf{Spatial}\\\textbf{Orientation}}
& \cellcolor{taskgreen}\makecell{\textbf{Mental}\\\textbf{Rotation}}
& \cellcolor{taskgreen}\makecell{\textbf{Spatial}\\\textbf{Visualization}} \\
\midrule
Metrics
& \makecell{Rep.\\Rate $\uparrow$}
& \makecell{Succ.\\Rate $\uparrow$}
& \makecell{Eff.\\Turn $\uparrow$}
& \makecell{Pre.\\Stop $\downarrow$}
& \makecell{Tok./\\Scene}
& Acc. $\uparrow$
& Acc. $\uparrow$
& Acc. $\uparrow$
& Acc. $\uparrow$
& Acc. $\uparrow$
& Acc. $\uparrow$ \\
\cmidrule(lr){1-12}

\rowcolor{gray!10}
\multicolumn{12}{l}{\textit{Proprietary Models}} \\
\midrule

GLM-5V-Turbo 
& -0.012 & 0.035 & 0.262 & 0.864 & 31101
& 0.452
& 0.695 & 0.738 & 0.617 & 0.500 & 0.665 \\

GPT-5.4 mini 
& 0.088 & 0.009 & 0.176 & 0.885 & 10060
& 0.595
& 0.720 & 0.810 &0.616& 0.570&0.813\\

GPT-5.4 
& 0.208 & 0.038 & 0.228 & 0.791 & 23040
& 0.664
& 0.751 & 0.826 & 0.625 & \textbf{0.690} & \textbf{0.850}\\

Gemini-3.1 Pro 
& \textbf{0.411} & \textbf{0.206} & \textbf{0.293} & \textbf{0.566} & 22814
& \textbf{0.721}
& \textbf{0.770} & \textbf{0.835} & \textbf{0.720} & 0.628 & 0.785 \\

\midrule
\rowcolor{gray!10}
\multicolumn{12}{l}{\textit{Open-source Models}} \\
\midrule

Kimi-K2.5 
& 0.032 & 0.009 & 0.055 & 0.920 & 9211
& 0.368
& 0.615 & 0.670 & 0.393 & 0.328 & 0.473 \\

Qwen3.6-27B 
& 0.035 & 0.005 & 0.020 & 0.922 & 8565
& 0.343
& 0.600 & 0.668 & 0.446 & 0.155 & 0.581 \\

Qwen3.6-35B-A3B 
& -0.099 & 0.016 & 0.252 & 0.920 & 196313
& 0.442
& 0.640 & 0.788 & 0.681 & 0.585 & 0.700 \\

\bottomrule
\end{tabular}

}
\vspace{-15pt}
\end{table*}

\vspace{-10pt}
\paragraph{Metric Design}
For the first-level Basic Spatial Abilities tasks and the second-level Single-step Error Detection and Fix task, we use accuracy as the evaluation metric.
For the Multi-turn Interactive Refinement task, we design five metrics targeting two aspects of model performance: repair accuracy and repair efficiency. We first introduce two metrics related to repair accuracy. First, we define the Repair Rate to measure the model's ability to reduce scene errors:
\vspace{-5pt}
\begin{equation}
\mathrm{Repair\ Rate} = 
\frac{E_{\mathrm{before}} - E_{\mathrm{after}}}{E_{\mathrm{before}}},
\end{equation}
where $E_{\mathrm{before}}$ and $E_{\mathrm{after}}$ denote the number of errors in the scenes before and after model refinement, respectively. This metric directly reflects how effectively the model repairs abnormal spatial configurations.

Second, we define the Scene Success Rate to evaluate whether the model can completely resolve all errors at the scene level:
\begin{equation}
\mathrm{Scene\ Success\ Rate} = 
\frac{1}{|\mathcal{S}|} \sum_{s \in \mathcal{S}} 
\mathbb{I}(E^{s}_{\mathrm{after}} = 0),
\end{equation}
\vspace{-2pt}
where $E^{s}_{\mathrm{after}}$ denotes the number of remaining errors in scene s after refinement. This metric measures the proportion of scenes that are fully corrected after model interaction, providing a scene-level evaluation of both error identification and correction.

Next, we introduce three efficiency-oriented metrics that measure how efficiently the model performs refinement during multi-turn interaction.
First, we define the Effective Repair Turn Ratio to measure the proportion of interaction turns that actually reduce scene errors:
\vspace{-2pt}
\begin{equation}
\mathrm{Effective\ Repair\ Turn\ Ratio} =
\frac{\sum_{s \in \mathcal{S}} \sum_{t=1}^{T_s}
\mathbb{I}(E^{s}_{t} < E^{s}_{t-1})}
{\sum_{s \in \mathcal{S}} T_s},
\end{equation}
where $E^{s}_{t}$ denotes the number of errors remaining in scene $s$ after the $t$-th refinement turn, and $T_s$ denotes the total number of interaction turns performed for scene $s$. This metric reflects how frequently the model produces effective actions during refinement. 

Second, we define the Premature Stop Rate to measure the proportion of scenes in which the model stops refinement while errors still remain:
\vspace{-2pt}
\begin{equation}
\mathrm{Premature\ Stop\ Rate} =
\frac{1}{|\mathcal{S}|}
\sum_{s \in \mathcal{S}}
\mathbb{I}(E^{s}_{T_s} > 0),
\end{equation}
\vspace{-2pt}
where $E^{s}_{T_s}$ denotes the number of remaining errors in scene $s$ at the end of the interaction, either when the model decides to stop or when the maximum number of turns is reached. This metric captures whether the model can correctly judge the completion status of the scene. 

Third, we define the Average Completion Tokens per Scene to measure the token cost:
\vspace{-2pt}
\begin{equation}
\mathrm{Avg.\ Completion\ Tokens} =
\frac{1}{|\mathcal{S}|}
\sum_{s \in \mathcal{S}} C_.
\end{equation}
\vspace{-2pt}
where $C_s$ denotes the total number of completion tokens generated by the model across all interaction turns for scene $s$. This metric evaluates the reasoning cost of multi-turn refinement. 

\vspace{-5pt}
\section{Experiments} \label{sec:exp}
\vspace{-5pt}

\subsection{Benchmark VLM Agents}
\vspace{-5pt}
We comprehensively evaluate 7 strong VLMs, covering diverse model scales and model families. For proprietary models, we include GPT-5.4 mini, GPT-5.4~\cite{gpt5.4}, GLM-5V-Turbo~\cite{hong2026glm}, and Gemini-3.1 Pro~\cite{gemini3.1-pro}. For open-source models, we evaluate Kimi-K2.5~\cite{kimiteam2026kimik25visualagentic}, Qwen3.6-27B~\cite{qwen3.6-27b}, and Qwen3.6-35B-A3B~\cite{qwen36_35b_a3b}.
During evaluation, we adopt a unified inference setting across all models. For the Multi-turn Interactive Refinement task, we set the maximum number of iterative turns to 30. Since the evaluated models are reasoning models with a context window of approximately 200K tokens, we limit the maximum number of completion tokens for each single-turn response to 8,096.
\vspace{-10pt}
\subsection{Main Results}
\begin{figure}[t]
    \centering
    \includegraphics[width=0.99\linewidth]{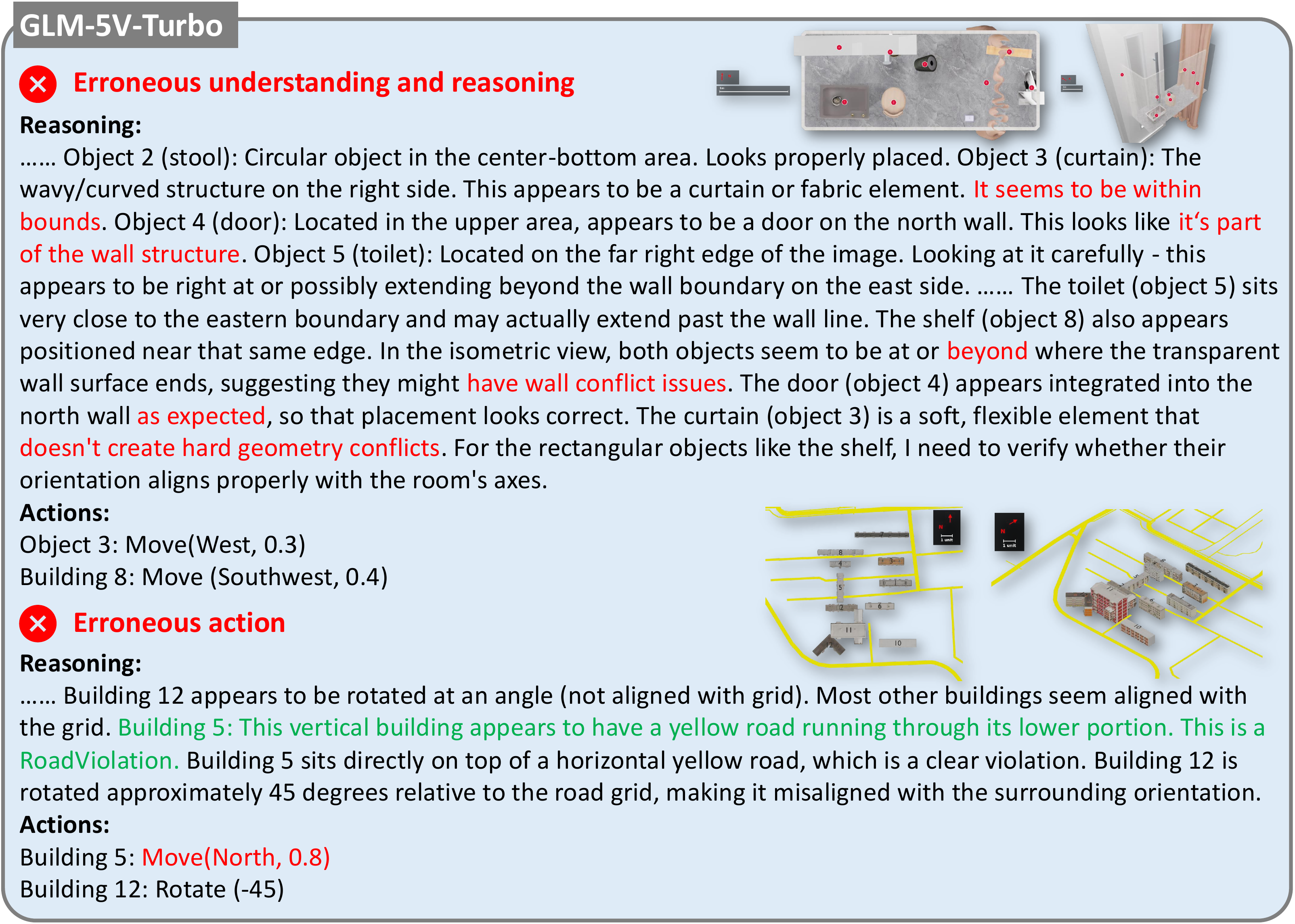}
    \caption{Representative error cases of GLM-5V-Turbo in SpatialAct.}
    \label{fig:case_study}
    \vspace{-15pt}
\end{figure}

\begin{figure}[t]
    \centering

    \begin{subfigure}[t]{0.33\linewidth}
        \centering
        \includegraphics[width=\linewidth]{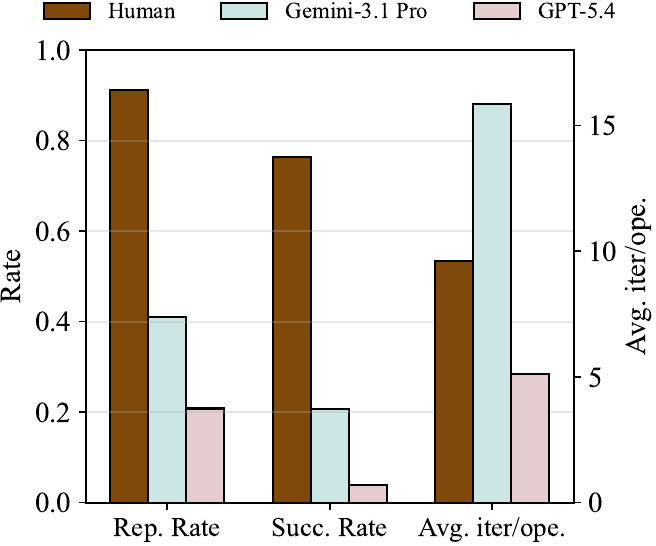}
        \vspace{-15pt}
        \caption{}
        \vspace{-5pt}
        \label{fig:human}
    \end{subfigure}
    \hfill
    \begin{subfigure}[t]{0.28\linewidth}
        \centering
        \includegraphics[width=\linewidth]{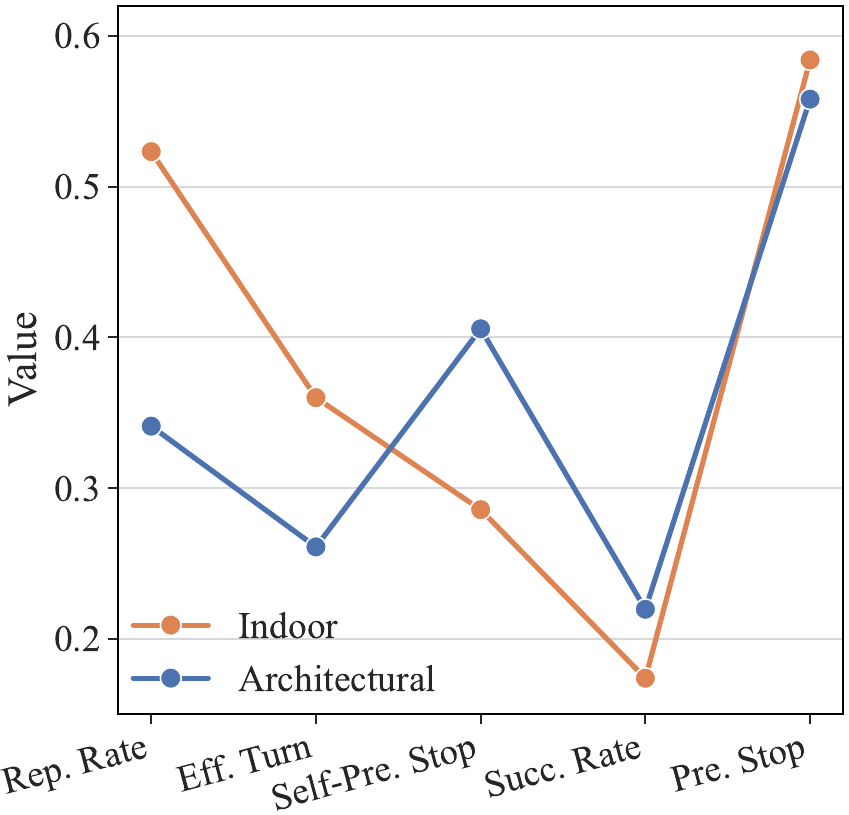}
        \vspace{-15pt}
        \caption{}
        \vspace{-5pt}
        \label{fig:indoor_region}
    \end{subfigure}%
    \hfill
    \begin{subfigure}[t]{0.30\linewidth}
        \centering
        \includegraphics[width=\linewidth]{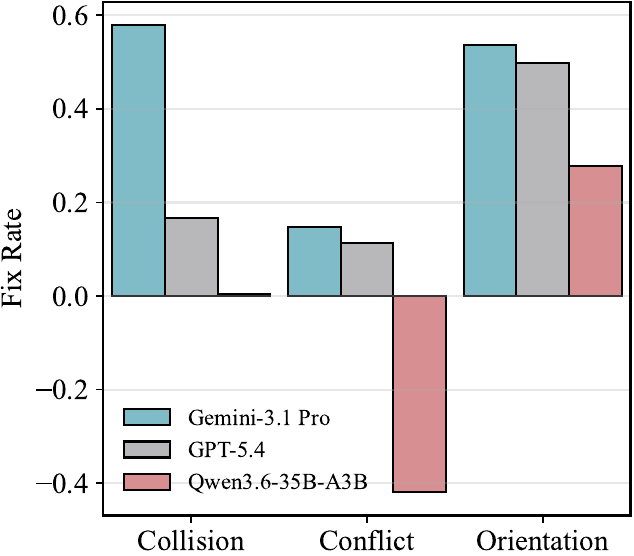}
        \vspace{-15pt}
        \caption{}
        \vspace{-5pt}
        \label{fig:specific_error_type}
    \end{subfigure}
    \caption{(a) Comparison of human and model performance. `Avg. iter./ope.' denotes average iteration/operation per scene. (b) Performance of the model across different layout types. `Self-Pre. Stop' denotes the fraction of turns in which the model chooses to stop before reaching the maximum allowed iterations while errors remain; lower values indicate better performance.
(c) Model sensitivity to different error types and correction performance.}
    \label{fig:first_three_column}
    \vspace{-10pt}
\end{figure}

\vspace{-5pt}
\paragraph{Overall Performance of Multi-turn Interactive Refinement}
As shown in Table~\ref{tab:main_results}, closed-source models substantially outperform other models on the Multi-turn Interactive Refinement task. In particular, Gemini-3.1 Pro achieve the leading performance, with Repair Rates of 0.411 and Scene Success Rates of 0.206, respectively. It also exhibits the highest refinement efficiency, indicating that stronger proprietary VLMs are not only more capable of reducing spatial errors, but also more effective in completing repairs with fewer redundant interactions and lower interaction cost.
In contrast, open-source models, as well as GLM-5V-Turbo, perform poorly on this task. Their Repair Rates are close to zero or even negative, suggesting that these models often fail to correct existing errors and may introduce additional errors during interaction. 
To make this bottleneck concrete, Figure~\ref{fig:case_study} shows two representative failure modes: the model can misdiagnose whether a spatial violation exists, and even when the diagnosis is correct, it can still issue an incorrect corrective action. Together, these cases indicate that the core weakness lies in both error diagnosis and reasoning-to-action execution.
Overall, Multi-turn Interactive Refinement remains a highly challenging task for existing VLMs. Although the strongest closed-source models show encouraging progress, their scene-level success rates are still far from saturation, while most open-source models struggle to perform reliable spatial repair. 
\vspace{-10pt}
\paragraph{Results on Single-step and Basic Spatial Ability Tasks}
We further evaluate model results on the Single-step Detection and Fix task and basic spatial ability tasks. Compared with Multi-turn Interactive Refinement, models generally achieve higher accuracies in these settings, suggesting that current VLMs are more capable of solving isolated spatial reasoning problems than performing long-horizon interactive repair.
For Single-step Detection and Fix, proprietary models maintain a clear advantage. Gemini-3.1 Pro achieves the best accuracy of 0.721, followed by GPT-5.4 and GPT-5.4 mini with 0.664 and 0.595, respectively, while open-source models remain much lower, ranging from 0.343 to 0.442. For basic spatial ability tasks, Gemini-3.1 Pro and GPT-5.4 remain the leading models across all five tasks.
Open-source models show uneven performance across basic abilities. For example, Kimi-K2.5 performs relatively well on Spatial Relation, but remains weaker on Mental Rotation. Overall, these results show that models exhibit diverse basic spatial capabilities, but these capabilities alone do not directly indicate their performance in multi-turn refinement.
\vspace{-10pt}
\paragraph{Human Baseline}
Seven human testers performed the Multi-turn Interactive Refinement task using the evaluation platform described in Section~\ref{sec:methods}, as shown in Figure~\ref{fig:human}. These repair tasks are straightforward for humans, who achieve a Repair Rate of 0.911 and a Scene Success Rate of 0.763. The repair rate exceeds that of the best-performing model by 50 percentage points, highlighting the remaining gaps in current VLMs in reasoning about scene errors, planning corrective actions, and executing multi-turn interactions.

\begin{insightbox}{Key Findings: Multi-turn interactive spatial correction remains a major bottleneck.}
\begin{itemize}[leftmargin=0.7em,labelsep=0.6em,itemindent=0pt,listparindent=0pt,topsep=2pt,itemsep=2pt]
\item \textbf{State-tracking limitation:} Current VLMs can solve many static or single-step spatial tasks, but fail to maintain consistent spatial state across iterative action-feedback loops.

\item \textbf{Failure modes:} In Multi-turn Interactive Refinement, current VLMs fail for two coupled reasons: \emph{diagnosis errors} (understanding and reasoning) and \emph{reasoning-to-action errors}.

\item \textbf{Human-level gap:} Although iterative spatial correction is simple for humans, current VLMs remain about 55\% lower in Repair Accuracy, revealing a substantial reliability gap.
\end{itemize}
\end{insightbox}

\vspace{-5pt}
\subsection{Spatial Context and Error Sensitivity}
\vspace{-5pt}
\paragraph{Scene-wise Analysis of VLM Performance}
We analyze the performance of Gemini-3.1 Pro separately on indoor and urban architectural scenes, as shown in Figure~\ref{fig:indoor_region}. Compared with architectural layouts, the model achieves higher Repair Rate and Effective Repair Turn Ratio in indoor scenes, and its self-selected Premature Stop Ratio is lower. This indicates that the model can more easily identify and correct errors in indoor scenes, likely reflecting greater exposure to such environments in its training data. 
Interestingly, architectural scenes show higher Scene Success Rate despite weaker per-turn repair effectiveness.
A plausible interpretation is that, although architectural scenes are less familiar to the model, their layouts may be simpler and include fewer tightly coupled object dependencies, making full-scene completion easier once key errors are addressed.
Overall, the model appears stronger in indoor repair behavior, while final scene success is also shaped by scene-level structural simplicity and error coupling.

\paragraph{Error-type Sensitivity in VLM Refinement}
Figure~\ref{fig:specific_error_type} shows a consistent error-type hierarchy across Gemini-3.1 Pro, GPT-5.4, and Qwen3.6-35B-A3B: orientation errors are most recoverable, while conflict errors, whether related to roads or walls, are the most challenging. This pattern suggests that current VLMs are relatively strong at attribute-level adjustments but weaker at constraint-level reasoning that requires jointly modeling boundaries, topology, and multi-object relations. Importantly, conflict repair often requires coordinated updates rather than single-object edits, so one incorrect step can propagate new violations across turns. These findings imply that the bottleneck is not only perception, but constraint-aware planning and cross-turn consistency under coupled spatial dependencies.
\vspace{-10pt}
\paragraph{Impact of Initial Scene Complexity on Model Performance}
As shown in Figure~\ref{fig:before_error}, we group indoor scenes by initial complexity based on the number of errors: 1–3, 4–6, and 7 or more. Not surprisingly, as the initial scene complexity increases, both Repair Rate and Scene Success Rate gradually decrease. 
This pattern reflects the increasing difficulty of tracking multiple errors simultaneously, which often leads to missed detections or conflicting corrective actions. Multi-turn interactions in such complex scenes also demand stronger long-horizon consistency and error management capabilities. Current VLMs are prone to interference when handling multiple errors, particularly in scenarios involving spatial conflicts or complex dependency relationships, resulting in poor refinement performance as scene complexity rises.
\begin{insightbox}[colback=orange!6!white,colframe=orange!70!black]{Key Findings: Joint effects of scene context, structural constraints, and error complexity.}
\begin{itemize}[leftmargin=0.7em,labelsep=0.6em,itemindent=0pt,listparindent=0pt,topsep=2pt,itemsep=2pt]
\item \textbf{Scene-dependent behavior:} VLMs are more effective at repairing errors in indoor scenes.

\item \textbf{Constraint-level bottleneck:} Across models, orientation errors are easier to repair than conflict errors, indicating that the main limitation lies in constraint-aware, multi-object coordination.

\item \textbf{Complexity amplifies instability:} Current VLMs struggle to maintain cross-turn consistency under dense, coupled error conditions.
\end{itemize}
\end{insightbox}

\vspace{-10pt}
\subsection{Influence of Context and Task Relationship on VLM Behavior}
\begin{figure}[t]
    \centering

    \begin{subfigure}[t]{0.33\linewidth}
        \centering
        \includegraphics[width=\linewidth]{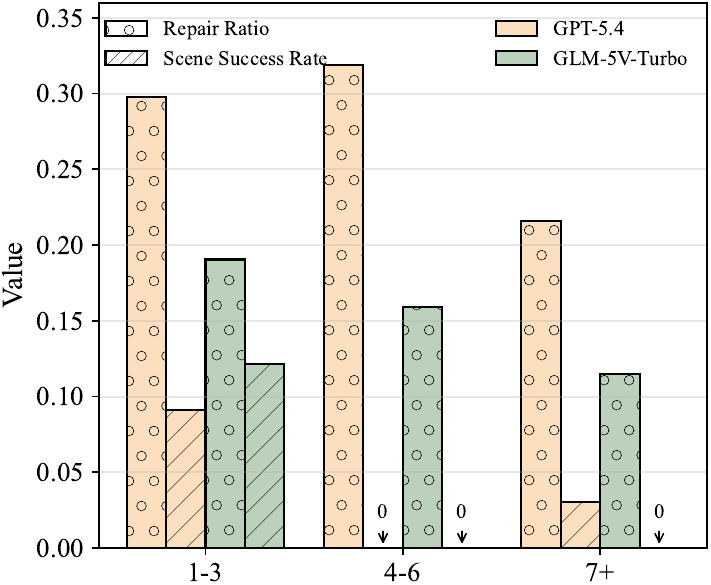}
        \vspace{-15pt}
        \caption{}
        \vspace{-5pt}
        \label{fig:before_error}
    \end{subfigure}
    \hfill
    \begin{subfigure}[t]{0.33\linewidth}
        \centering
        \includegraphics[width=\linewidth]{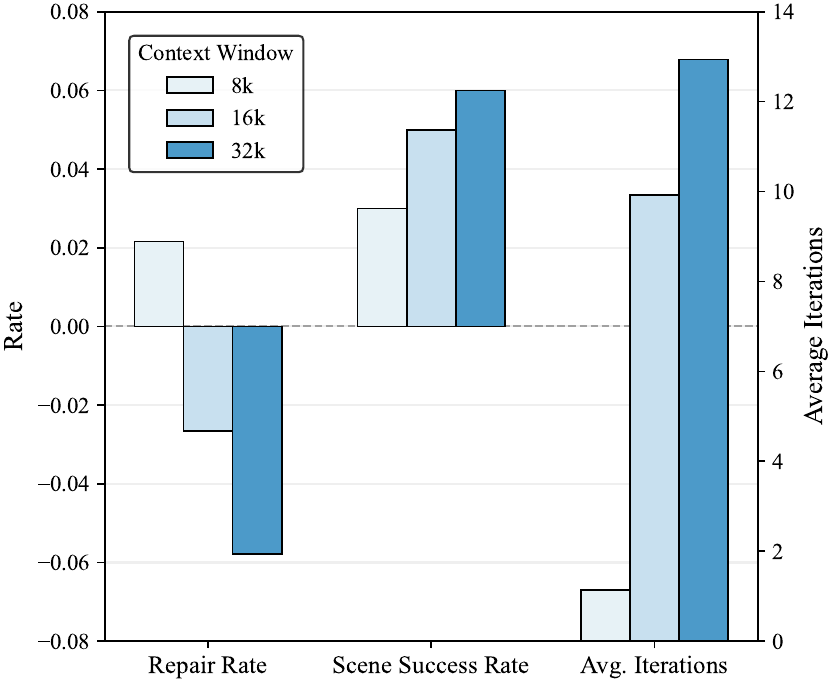}
        \vspace{-15pt}
        \caption{}
        \vspace{-5pt}
        \label{fig:context_window}
    \end{subfigure}%
    \hfill
    \begin{subfigure}[t]{0.30\linewidth}
        \centering
        \includegraphics[width=\linewidth]{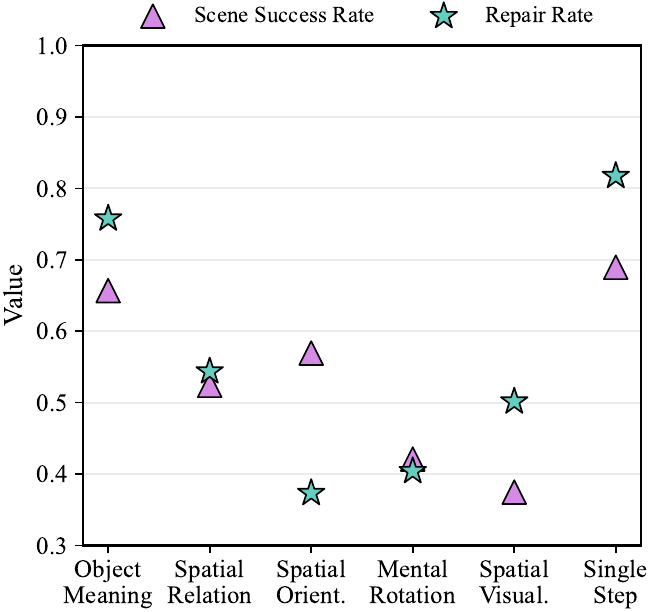}
        \vspace{-15pt}
        \caption{}
        \vspace{-5pt}
        \label{fig:correlation}
    \end{subfigure}
    \caption{(a) Comparison of model performance under different initial scene complexities. 
(b) Model performance under different context window settings. 
(c) Relationship between performance on basic spatial tasks and complex interactive tasks, quantified by Pearson correlation coefficients ($r$).}
    \label{fig:second_three_column}
     \vspace{-10pt}
\end{figure}

\vspace{-5pt}
\paragraph{Effect of Context Window Size}
As illustrated in Figure~\ref{fig:context_window}, we analyze context-window effects on Kimi-K2.5, a model that tends to produce long reasoning traces, using a random subset of 100 test examples. As the context window increases, the model produces more environment-applicable content and engages in more interaction turns, indicating that additional context is converted into more explicit deliberation and action attempts. However, Repair Rate and Scene Success Rate remain nearly unchanged. This decoupling between \emph{reasoning volume} and \emph{repair outcome} suggests a diminishing-return regime in which longer contexts primarily expand verbose or repetitive reasoning rather than improving correction quality. A likely explanation is that the bottleneck is not token budget itself, but cross-turn control quality, including state tracking, error prioritization, and action reliability under feedback. Under this view, larger windows increase opportunity for exploration but do not strengthen the policy that selects effective repairs. This also supports the context setting used in Table~\ref{tab:main_results}: once a sufficient reasoning budget is reached, further context expansion adds computational cost with limited performance gain.
\vspace{-10pt}
\paragraph{Correlation between Simple and Complex Task Performance}
Figure~\ref{fig:correlation} relates performance on six foundational tasks (five Basic Spatial Abilities plus Single-step Error Detection and Fix) to Multi-turn Interactive Refinement. Positive Pearson correlations across all six tasks indicate structural continuity between task levels, suggesting that foundational abilities provide the operational components required in complex multi-turn repair. Single-step Error Detection and Fix shows the strongest association with both Repair Rate ($r=0.817$) and Scene Success Rate ($r=0.690$), indicating that local detect-and-correct operations serve as the fundamental building blocks repeatedly composed in iterative refinement. Object Meaning is the second strongest correlate, suggesting that object-centric grounding, including identity, position, and relative spatial anchoring, is critical for propagating local edits to scene-level consistency.
At the same time, the correlation pattern shows that foundational competence alone does not guarantee robust multi-turn success, since iterative repair additionally depends on cross-turn memory, conflict-aware planning, and stable action sequencing. Overall, complex refinement can be understood as a hierarchical integration of foundational spatial skills with higher-order coordination over feedback loops.
\begin{insightbox}[colback=cyan!6!white,colframe=cyan!65!black]{\textbf{Key Findings: Control quality and hierarchical skill integration over context length.}}
\begin{itemize}[leftmargin=1em,itemsep=2pt,topsep=2pt]
\item \textbf{Diminishing returns from context scaling:} More tokens and turns, but limited improvement in Repair Rate and Scene Success Rate.
\item \textbf{Hierarchical but non-automatic skill transfer:} Foundational abilities align with multi-turn demands, yet robust performance still requires higher-order cross-turn coordination.
\end{itemize}
\end{insightbox}

 \vspace{-10pt}

\section{Related Work} \label{sec:related}
\vspace{-5pt}
\paragraph{3D Spatial Intelligence Evaluation}
With the rapid development of VLMs, increasing attention has been paid to whether these models can perceive, reason about, and act upon spatial information in a human-like manner. This has led to a growing body of benchmarks for evaluating spatial intelligence from different perspectives. Some benchmarks~\cite{wang2025spatial457,zhang-etal-2025-sphere,wang2025site,ramakrishnan2024does,kanade2026yousee,liu2025spatialsurvey,yu2025far,xiao2026spatialtree,rahmanzadehgervi2024vision}, such as Spatial-DISE~\cite{huang2025spatial} and BSA~\cite{xu2025defining}, construct hierarchical suites of classical spatial tasks to assess fundamental spatial abilities in a systematic way. Recent work has further expanded the scope of spatial evaluation to more diverse and realistic scenarios~\cite{wang2024picture,li2024core}. For example, SpaceVista~\cite{sun2025spacevista} investigates spatial understanding across all-scale scenes ranging from millimeters to kilometers, while other benchmarks extend object-centric reasoning from single-step judgement to multi-step interaction with the environment~\cite{zhang2026theory}. Meanwhile, model inputs have also evolved from single-view observations to richer multi-view settings~\cite{yang2025thinking,yin2025spatial,zhao2025cityeqa,li2025viewspatial}. CityCube~\cite{xu2026citycube}, for instance, collects images from different viewing positions and orientations to support more comprehensive spatial perception.
In parallel, the objective of spatial evaluation has gradually shifted from passive spatial understanding to action-oriented spatial interaction~\cite{wang2025embodied,zhang2025etplan,limanling2024embodied}, where models are expected not only to recognize spatial relations but also to modify the environment through actions. Our SpatialAct follows this direction and provides a comprehensive benchmark for evaluating agentic VLMs in 3D environments from reasoning to action. 
\vspace{-10pt}
\paragraph{3D Layout Generation and Understanding}
Recent advances in AI for 3D layout have explored both layout generation~\cite{feng2026repurposing} and layout understanding~\cite{huang2025fireplace,abdelreheem2025placeit3d,el2025scanedit}. On the generation side, existing methods increasingly leverage foundation models to inject semantic and spatial commonsense into 3D layout synthesis. For example, LayoutGPT~\cite{feng2023layoutgpt} directly uses LLMs to produce structured layout representations, while Holodeck~\cite{yang2024holodeck} and LayoutVLM~\cite{sun2025layoutvlm} further combine VLM reasoning with spatial constraints or differentiable optimization to improve semantic coherence and physical plausibility.
On the understanding side, recent work has begun to move beyond passive spatial perception toward layout editing and manipulation. \cite{weihs2021rearrangement} studies embodied agents that restore shuffled indoor scenes through observation and interaction, while 3D-Layout-R1~\cite{zhen20263d-layout-r1} formulates language-guided layout editing as structured scene-graph reasoning.
However, most existing 3D layout works still leave open the question of whether vision-language models can directly complete layout tasks through reasoning and action. Since 3D layout quality critically depends on physical validity~\cite{sun2025layoutvlm,feng2026repurposing}, such as avoiding collisions, out-of-bound placements, and spatial misalignment, We introduce SpatialAct to evaluate whether agentic VLMs can reason over interactive 3D environments and generate action commands to iteratively repair or modify spatial layouts.

\vspace{-5pt}
\section{Conclusion}\label{sec:conclusion}
\vspace{-5pt}
We present SpatialAct, a simulator-grounded hierarchical benchmark that probes whether VLMs can translate spatial reasoning into reliable action under dynamic 3D feedback. By jointly evaluating Multi-turn Interactive Refinement, Single-step Error Detection and Fix, and basic spatial abilities, SpatialAct reveals not only absolute performance gaps but also the capability structure behind them. Across models, strong results on basic tasks do not consistently transfer to robust multi-turn repair, indicating that the core bottleneck lies in cross-turn state maintenance, constraint-aware planning, and stable reasoning-to-action execution rather than perception alone. Proprietary models remain clearly ahead of open-source models, but all current systems still fall far short of human reliability on iterative spatial correction. We further find that performance is systematically shaped by scene context, structural constraints, and error complexity, with models generally handling indoor scenes and orientation-related corrections more effectively. Beyond benchmarking, SpatialAct provides an actionable diagnostic framework for developing VLM agents with stronger spatial state tracking, coordination across feedback loops, and dependable context-grounded action generation.

\newpage
\bibliographystyle{plain}
\bibliography{reference}

\begin{thebibliography}{10}

\bibitem{abdelreheem2025placeit3d}
Ahmed Abdelreheem, Filippo Aleotti, Jamie Watson, Zawar Qureshi, Abdelrahman Eldesokey, Peter Wonka, Gabriel Brostow, Sara Vicente, and Guillermo Garcia-Hernando.
\newblock Placeit3d: Language-guided object placement in real 3d scenes.
\newblock In {\em Proceedings of the IEEE/CVF International Conference on Computer Vision}, pages 6645--6655, 2025.

\bibitem{gemini3.1-pro}
Google DeepMind.
\newblock Gemini-3.1 pro.
\newblock \url{https://deepmind.google/models/model-cards/gemini-3-1-pro/}, 2026.

\bibitem{el2025scanedit}
Mohamed El~Amine~Boudjoghra, Ivan Laptev, and Angela Dai.
\newblock Scanedit: Hierarchically-guided functional 3d scan editing.
\newblock In {\em Proceedings of the IEEE/CVF International Conference on Computer Vision}, pages 27105--27115, 2025.

\bibitem{feng2026repurposing}
Haoran Feng, Yifan Niu, Zehuan Huang, Yang-Tian Sun, Chunchao Guo, Yuxin Peng, and Lu~Sheng.
\newblock Repurposing 3d generative model for autoregressive layout generation.
\newblock {\em arXiv preprint arXiv:2604.16299}, 2026.

\bibitem{feng2023layoutgpt}
Weixi Feng, Wanrong Zhu, Tsu-jui Fu, Varun Jampani, Arjun Akula, Xuehai He, Sugato Basu, Xin~Eric Wang, and William~Yang Wang.
\newblock Layoutgpt: Compositional visual planning and generation with large language models.
\newblock {\em Advances in Neural Information Processing Systems}, 36:18225--18250, 2023.

\bibitem{hong2026glm}
Wenyi Hong, Xiaotao Gu, Ziyang Pan, Zhen Yang, Yuting Wang, Yue Wang, Yuanchang Yue, Yu~Wang, Yanling Wang, Yan Wang, et~al.
\newblock Glm-5v-turbo: Toward a native foundation model for multimodal agents.
\newblock {\em arXiv preprint arXiv:2604.26752}, 2026.

\bibitem{huang2025fireplace}
Ian Huang, Yanan Bao, Karen Truong, Howard Zhou, Cordelia Schmid, Leonidas Guibas, and Alireza Fathi.
\newblock Fireplace: Geometric refinements of llm common sense reasoning for 3d object placement.
\newblock In {\em Proceedings of the IEEE/CVF conference on computer vision and pattern recognition}, pages 13466--13476, 2025.

\bibitem{huang2025spatial}
Xinmiao Huang, Qisong He, Zhenglin Huang, Boxuan Wang, Zhuoyun Li, Guangliang Cheng, Yi~Dong, and Xiaowei Huang.
\newblock Spatial-dise: A unified benchmark for evaluating spatial reasoning in vision-language models.
\newblock {\em arXiv preprint arXiv:2510.13394}, 2025.

\bibitem{kanade2026yousee}
Aditya~Sanjiv Kanade and Tanuja Ganu.
\newblock Do you see me: A multidimensional benchmark for evaluating visual perception in multimodal llms.
\newblock In {\em Proceedings of the 19th Conference of the European Chapter of the Association for Computational Linguistics (Volume 1: Long Papers)}, pages 7285--7326, 2026.

\bibitem{li2025viewspatial}
Dingming Li, Hongxing Li, Zixuan Wang, Yuchen Yan, Hang Zhang, Siqi Chen, Guiyang Hou, Shengpei Jiang, Wenqi Zhang, Yongliang Shen, et~al.
\newblock Viewspatial-bench: Evaluating multi-perspective spatial localization in vision-language models.
\newblock {\em arXiv preprint arXiv:2505.21500}, 2025.

\bibitem{limanling2024embodied}
Manling Li, Shiyu Zhao, Qineng Wang, Kangrui Wang, Yu~Zhou, Sanjana Srivastava, Cem Gokmen, Tony Lee, Li~Li, Ruohan Zhang, Weiyu Liu, Percy Liang, Li~Fei-Fei, Jiayuan Mao, and Jiajun Wu.
\newblock Embodied agent interface: Benchmarking llms for embodied decision making.
\newblock In A.~Globerson, L.~Mackey, D.~Belgrave, A.~Fan, U.~Paquet, J.~Tomczak, and C.~Zhang, editors, {\em Advances in Neural Information Processing Systems}, volume~37, pages 100428--100534. Curran Associates, Inc., 2024.

\bibitem{li2024core}
Yijiang Li, Qingying Gao, Tianwei Zhao, Bingyang Wang, Haoran Sun, Haiyun Lyu, Robert~D Hawkins, Nuno Vasconcelos, Tal Golan, Dezhi Luo, et~al.
\newblock Core knowledge deficits in multi-modal language models.
\newblock {\em arXiv preprint arXiv:2410.10855}, 2024.

\bibitem{liu2025spatialsurvey}
Weichen Liu, Qiyao Xue, Haoming Wang, Xiangyu Yin, Boyuan Yang, and Wei Gao.
\newblock Spatial reasoning in multimodal large language models: A survey of tasks, benchmarks and methods.
\newblock {\em arXiv preprint arXiv:2511.15722}, 2025.

\bibitem{majumdar2024openeqa}
Arjun Majumdar, Anurag Ajay, Xiaohan Zhang, Pranav Putta, Sriram Yenamandra, Mikael Henaff, Sneha Silwal, Paul Mcvay, Oleksandr Maksymets, Sergio Arnaud, et~al.
\newblock Openeqa: Embodied question answering in the era of foundation models.
\newblock In {\em Proceedings of the IEEE/CVF conference on computer vision and pattern recognition}, pages 16488--16498, 2024.

\bibitem{gpt5.4}
OpenAI.
\newblock Gpt-5.4.
\newblock \url{https://openai.com/index/introducing-gpt-5-4/}, 2026.

\bibitem{qwen3.6-27b}
{Qwen Team}.
\newblock {Qwen3.6-27B}: Flagship-level coding in a {27B} dense model, April 2026.

\bibitem{qwen36_35b_a3b}
{Qwen Team}.
\newblock {Qwen3.6-35B-A3B}: Agentic coding power, now open to all, April 2026.

\bibitem{rahmanzadehgervi2024vision}
Pooyan Rahmanzadehgervi, Logan Bolton, Mohammad~Reza Taesiri, and Anh~Totti Nguyen.
\newblock Vision language models are blind: Failing to translate detailed visual features into words.
\newblock {\em arXiv preprint arXiv:2407.06581}, 2024.

\bibitem{ramakrishnan2024does}
Santhosh~Kumar Ramakrishnan, Erik Wijmans, Philipp Kraehenbuehl, and Vladlen Koltun.
\newblock Does spatial cognition emerge in frontier models?
\newblock {\em arXiv preprint arXiv:2410.06468}, 2024.

\bibitem{sun2025layoutvlm}
Fan-Yun Sun, Weiyu Liu, Siyi Gu, Dylan Lim, Goutam Bhat, Federico Tombari, Manling Li, Nick Haber, and Jiajun Wu.
\newblock Layoutvlm: Differentiable optimization of 3d layout via vision-language models.
\newblock In {\em Proceedings of the Computer Vision and Pattern Recognition Conference}, pages 29469--29478, 2025.

\bibitem{sun2025spacevista}
Peiwen Sun, Shiqiang Lang, Dongming Wu, Yi~Ding, Kaituo Feng, Huadai Liu, Zhen Ye, Rui Liu, Yun-Hui Liu, Jianan Wang, et~al.
\newblock Spacevista: All-scale visual spatial reasoning from mm to km.
\newblock {\em arXiv preprint arXiv:2510.09606}, 2025.

\bibitem{kimiteam2026kimik25visualagentic}
Kimi Team and et~al.
\newblock Kimi k2.5: Visual agentic intelligence, 2026.

\bibitem{wang2024picture}
Jiayu Wang, Yifei Ming, Zhenmei Shi, Vibhav Vineet, Xin Wang, Yixuan Li, and Neel Joshi.
\newblock Is a picture worth a thousand words? delving into spatial reasoning for vision language models.
\newblock {\em Advances in Neural Information Processing Systems}, 37:75392--75421, 2024.

\bibitem{wang2025raisecity}
Shengyuan Wang, Zhiheng Zheng, Yu~Shang, Lixuan He, Yangcheng Yu, Fan Hangyu, Jie Feng, Qingmin Liao, and Yong Li.
\newblock Raisecity: A multimodal agent framework for reality-aligned 3d world generation at city-scale.
\newblock {\em arXiv preprint arXiv:2511.18005}, 2025.

\bibitem{wang2025embodied}
Weizhen Wang, Chenda Duan, Zhenghao Peng, Yuxin Liu, and Bolei Zhou.
\newblock Embodied scene understanding for vision language models via metavqa.
\newblock In {\em Proceedings of the IEEE/CVF Conference on Computer Vision and Pattern Recognition}, pages 22453--22464, 2025.

\bibitem{wang2025site}
Wenqi Wang, Reuben Tan, Pengyue Zhu, Jianwei Yang, Zhengyuan Yang, Lijuan Wang, Andrey Kolobov, Jianfeng Gao, and Boqing Gong.
\newblock Site: towards spatial intelligence thorough evaluation.
\newblock In {\em Proceedings of the IEEE/CVF International Conference on Computer Vision}, pages 9058--9069, 2025.

\bibitem{wang2025spatial457}
Xingrui Wang, Wufei Ma, Tiezheng Zhang, Celso~M de~Melo, Jieneng Chen, and Alan Yuille.
\newblock Spatial457: A diagnostic benchmark for 6d spatial reasoning of large mutimodal models.
\newblock In {\em Proceedings of the Computer Vision and Pattern Recognition Conference}, pages 24669--24679, 2025.

\bibitem{weihs2021rearrangement}
Luca Weihs, Matt Deitke, Aniruddha Kembhavi, and Roozbeh Mottaghi.
\newblock Visual room rearrangement.
\newblock In {\em Proceedings of the IEEE/CVF conference on computer vision and pattern recognition}, pages 5922--5931, 2021.

\bibitem{xiao2026spatialtree}
Yuxi Xiao, Longfei Li, Shen Yan, Xinhang Liu, Sida Peng, Yunchao Wei, Xiaowei Zhou, and Bingyi Kang.
\newblock Spatialtree : How spatial abilities branch out in {MLLM}s.
\newblock In {\em The First Workshop on Efficient Spatial Reasoning}, 2026.

\bibitem{xu2026citycube}
Haotian Xu, Yue Hu, Zhengqiu Zhu, Chen Gao, Ziyou Wang, Junreng Rao, Wenhao Lu, Weishi Li, Quanjun Yin, and Yong Li.
\newblock Citycube: Benchmarking cross-view spatial reasoning on vision-language models in urban environments.
\newblock {\em arXiv preprint arXiv:2601.14339}, 2026.

\bibitem{xu2025defining}
Wenrui Xu, Dalin Lyu, Weihang Wang, Jie Feng, Chen Gao, and Yong Li.
\newblock Defining and evaluating visual language models’ basic spatial abilities: A perspective from psychometrics.
\newblock In {\em Proceedings of the 63rd Annual Meeting of the Association for Computational Linguistics (Volume 1: Long Papers)}, pages 11571--11590, 2025.

\bibitem{yang2025thinking}
Jihan Yang, Shusheng Yang, Anjali~W Gupta, Rilyn Han, Li~Fei-Fei, and Saining Xie.
\newblock Thinking in space: How multimodal large language models see, remember, and recall spaces.
\newblock In {\em Proceedings of the Computer Vision and Pattern Recognition Conference}, pages 10632--10643, 2025.

\bibitem{yang2024holodeck}
Yue Yang, Fan-Yun Sun, Luca Weihs, Eli VanderBilt, Alvaro Herrasti, Winson Han, Jiajun Wu, Nick Haber, Ranjay Krishna, Lingjie Liu, et~al.
\newblock Holodeck: Language guided generation of 3d embodied ai environments.
\newblock In {\em Proceedings of the IEEE/CVF Conference on Computer Vision and Pattern Recognition}, pages 16227--16237, 2024.

\bibitem{yin2025spatial}
Baiqiao Yin, Qineng Wang, Pingyue Zhang, Jianshu Zhang, Kangrui Wang, Zihan Wang, Jieyu Zhang, Keshigeyan Chandrasegaran, Han Liu, Ranjay Krishna, et~al.
\newblock Spatial mental modeling from limited views.
\newblock In {\em Structural Priors for Vision Workshop at ICCV'25}, 2025.

\bibitem{yu2025far}
Songsong Yu, Yuxin Chen, Hao Ju, Lianjie Jia, Fuxi Zhang, Shaofei Huang, Yuhan Wu, Rundi Cui, Binghao Ran, Zaibin Zhang, et~al.
\newblock How far are vlms from visual spatial intelligence? a benchmark-driven perspective.
\newblock {\em arXiv preprint arXiv:2509.18905}, 2025.

\bibitem{zhang2025etplan}
Lingfeng Zhang, Yuening Wang, Hongjian Gu, Atia Hamidizadeh, Zhanguang Zhang, Yuecheng Liu, Yutong Wang, David Gamaliel~Arcos Bravo, Junyi Dong, Shunbo Zhou, et~al.
\newblock Et-plan-bench: Embodied task-level planning benchmark towards spatial-temporal cognition with foundation models.
\newblock In {\em 2025 IEEE/RSJ International Conference on Intelligent Robots and Systems (IROS)}, pages 21566--21573. IEEE, 2025.

\bibitem{zhang2026theory}
Pingyue Zhang, Zihan Huang, Yue Wang, Jieyu Zhang, Letian Xue, Zihan Wang, Qineng Wang, Keshigeyan Chandrasegaran, Ruohan Zhang, Yejin Choi, et~al.
\newblock Theory of space: Can foundation models construct spatial beliefs through active exploration?
\newblock In {\em The Fourteenth International Conference on Learning Representations}, 2026.

\bibitem{zhang-etal-2025-sphere}
Wenyu Zhang, Wei~En Ng, Lixin Ma, Yuwen Wang, Junqi Zhao, Allison Koenecke, Boyang Li, and Lu~Wang.
\newblock {SPHERE}: Unveiling spatial blind spots in vision-language models through hierarchical evaluation.
\newblock In {\em Proceedings of the 63rd Annual Meeting of the Association for Computational Linguistics (Volume 1: Long Papers)}, pages 11591--11609, Vienna, Austria, July 2025. Association for Computational Linguistics.

\bibitem{zhao2025cityeqa}
Yong Zhao, Kai Xu, Zhengqiu Zhu, Yue Hu, Zhiheng Zheng, Yingfeng Chen, Yatai Ji, Chen Gao, Yong Li, and Jincai Huang.
\newblock Cityeqa: A hierarchical llm agent on embodied question answering benchmark in city space.
\newblock In {\em Proceedings of the 2025 Conference on Empirical Methods in Natural Language Processing}, pages 12476--12491, 2025.

\bibitem{zhen20263d-layout-r1}
Haoyu Zhen, Xiaolong Li, Yilin Zhao, Han Zhang, Sifei Liu, Kaichun Mo, Chuang Gan, and Subhashree Radhakrishnan.
\newblock 3d-layout-r1: Structured reasoning for language-instructed spatial editing.
\newblock {\em arXiv preprint arXiv:2603.22279}, 2026.

\bibitem{zhong2025internscenes}
Weipeng Zhong, Peizhou Cao, Yichen Jin, Li~Luo, Wenzhe Cai, Jingli Lin, Hanqing Wang, Zhaoyang Lyu, Tai Wang, Bo~Dai, et~al.
\newblock Internscenes: A large-scale simulatable indoor scene dataset with realistic layouts.
\newblock {\em arXiv preprint arXiv:2509.10813}, 2025.

\end{thebibliography}

\newpage
\appendix
\section{Appendix}
\subsection{Discussion and Future Work}
Our results reveal persistent challenges in multi-turn reasoning and action planning, with models showing systematic biases across scenes and error types. A key limitation is that all evaluations are conducted in simulated scenes, and it remains unclear how performance translates to real-world environments. Additionally, we did not explore methods to enhance model capabilities for this task. Future work includes extending evaluation to real-world scenes, exploring methods to improve multi-turn reasoning and action execution, and developing more robust, human-like spatial reasoning-to-action capabilities.
\subsection{Task Subcategories and Prompt Design}
\label{app:subcate_bsa}
To provide a fine-grained diagnosis of model performance, we further divide the Basic Spatial Ability tasks and the Single-step Error Detection and Fix task into multiple subcategories. As shown in Table~\ref{tab:task_subcategories}, the five Basic Spatial Ability tasks cover Object Meaning, Spatial Relation, Spatial Orientation, Mental Rotation, and Spatial Visualization, with each category containing several subtypes that emphasize different aspects of 3D spatial understanding and reasoning. In addition, the Single-step Error Detection and Fix task is divided according to different error patterns, enabling targeted evaluation of whether models can identify abnormal spatial configurations and select the corresponding correction. Representative prompts for each subcategory are provided in the table.

\begin{table}[ht]
\centering
\footnotesize
\caption{Subcategories and representative prompts for Basic Spatial Ability tasks and Single-step Error Detection and Fix tasks.}
\label{tab:task_subcategories}
\setlength{\tabcolsep}{5pt} %
\begin{minipage}[t]{\linewidth}
\begin{tabular}{lp{0.65\linewidth}} 
\toprule
\textbf{Task}& \textbf{Prompt}  \\
\midrule
Object Meaning (7)& How many bins are in the room?\\
 &What is object 1?\\
 &Which numbered object is a commode? \\
 &Which side of the box is the frame on?\\
 &Which numbered object is east of the commode?\\
 &Which numbered object is closest to the cabinet?\\
 &Which numbered object is farthest from the cabinet?\\
\midrule
Spatial Relation (4)&Which side is building 1 relative to building 5?\\
 &Which two buildings are closest to each other?\\
 &Which building is farthest from building 2?\\
 &How many buildings are on the east side of building 3?\\
\midrule
Spatial Orientation (3)& Consider the southernmost building in the isometric image. Relative to the scene center, which direction is that building located in the top-view image?\\

 &An isometric image where only one building is labeled 'A', and a top-view image with numbered buildings. Which numbered building is 'A' in the top-view image?\\
 &Identify the westernmost building in the image. Now imagine that the camera rotates CLOCKWISE by 90 degrees around the scene center.When the camera rotates, the cardinal directions (north, east, south, west) rotate together with the view. After this rotation, where will that building be located relative to the scene center in the rotated coordinate frame?\\
 \midrule
 Mental Rotation (4)&If building 1 is rotated counterclockwise by 30 degrees around its own center, will it collide with building 4? \\
& Rotate building 1 counterclockwise by one of the angles below around its own center. Which rotation angle avoids collision with building 4?\\
& Building 1 is rotated clockwise by 300 degrees.Which image matches this rotation?\\
& In the original image, Building 6 is rotated clockwise by 270 degrees around the center of the region. Which option shows the correct result after this rotation?\\
\midrule
Spatial Visualization (2) &If building 1 is moved West by 6 meters, will it collide with building 6? \\
&If we swap the positions of building 4 and building 5, will there be any collision in the scene AFTER the swap?\\
\midrule
Single-step Error Detection and Fix (2)
& Examine the scene carefully and identify what problem exists.\\
& Known issue: Building 1 has an angle anomaly near a road intersection. Choose ONE action to fix it in ONE step. The fix should resolve the issue without introducing new problems.\\
 \bottomrule
\end{tabular}
\end{minipage}
\end{table}

\end{document}